\documentclass[conference]{IEEEtran}

\usepackage{times}
\usepackage[T1]{fontenc}
\usepackage[utf8]{inputenc}
\usepackage{textcomp}
\usepackage[protrusion=true,expansion=true]{microtype}

\usepackage[numbers]{natbib}
\usepackage{amsmath}
\usepackage{amssymb}
\usepackage{bm}
\usepackage{graphicx}
\usepackage[table]{xcolor}
\usepackage{colortbl}
\usepackage{booktabs}
\usepackage{multirow}
\usepackage{makecell}
\usepackage{pifont}
\usepackage{placeins}
\usepackage{tabularx}
\usepackage{array}
\usepackage{url}
\usepackage{bbding}
\usepackage[bookmarks=true,breaklinks=true,colorlinks=true,allcolors=blue]{hyperref}

\renewcommand{\arraystretch}{1.08}

\begin{document}

\title{${\mathcal{P}}^3$: Toward Versatile Embodied Agents}

\author{
\authorblockN{Shengli Zhou$^{1*}$, Xiangchen Wang$^{1*}$, Jinrui Zhang$^1$, Ruozai Tian$^2$, Rongtao Xu$^3$, Guanhua Chen$^{1\dagger}$, Feng Zheng$^{1,2\dagger}$}
\authorblockA{
$^1$Southern University of Science and Technology\ \ $^2$Spatialtemporal AI\ \ $^3$MBZUAI\\
{\tt\small \{zhousl2022, wangxc2019, zhangjr2018, 12110924\}@mail.sustech.edu.cn}\\
{\tt\small xurongtao2022@gmail.com, chengh3@sustech.edu.cn, f.zheng@ieee.org}\\
$^*$Equal Contribution \quad $^\dagger$Corresponding Author
}
}

\maketitle

\begin{abstract}
Embodied agents have shown promising generalization capabilities across diverse physical environments, making them essential for a wide range of real-world applications. However, building versatile embodied agents poses critical challenges due to three key issues: dynamic environment perception, open-ended tool usage, and complex multi-task planning. Most previous works rely solely on feedback from tool agents to perceive environmental changes and task status, which limits adaptability to real-time dynamics, causes error accumulation, and restricts tool flexibility. Furthermore, multi-task scheduling has received limited attention, primarily due to the inherent complexity of managing task dependencies and balancing competing priorities in dynamic and complex environments. To overcome these challenges, we introduce $\mathcal P^3$, a unified framework that integrates real-time perception and dynamic scheduling. Specifically, $\mathcal P^3$ enables agents to \textbf{Perceive} task-relevant information actively from the environment, \textbf{Plug} and utilize tools without feedback requirements, and \textbf{Plan} multi-task execution by prioritizing urgent tasks and dynamically adjusting task order based on dependencies. Extensive real-world experiments show that our approach bridges the gap between benchmarks and practical deployment, delivering highly transferable, general-purpose embodied agents. Code and data are available at \href{https://github.com/fz-zsl/P3}{https://github.com/fz-zsl/P3}.
\end{abstract}

\IEEEpeerreviewmaketitle

\section{Introduction}

\begin{figure*}[tbp]
    \centering
    \includegraphics[width=0.96\textwidth]{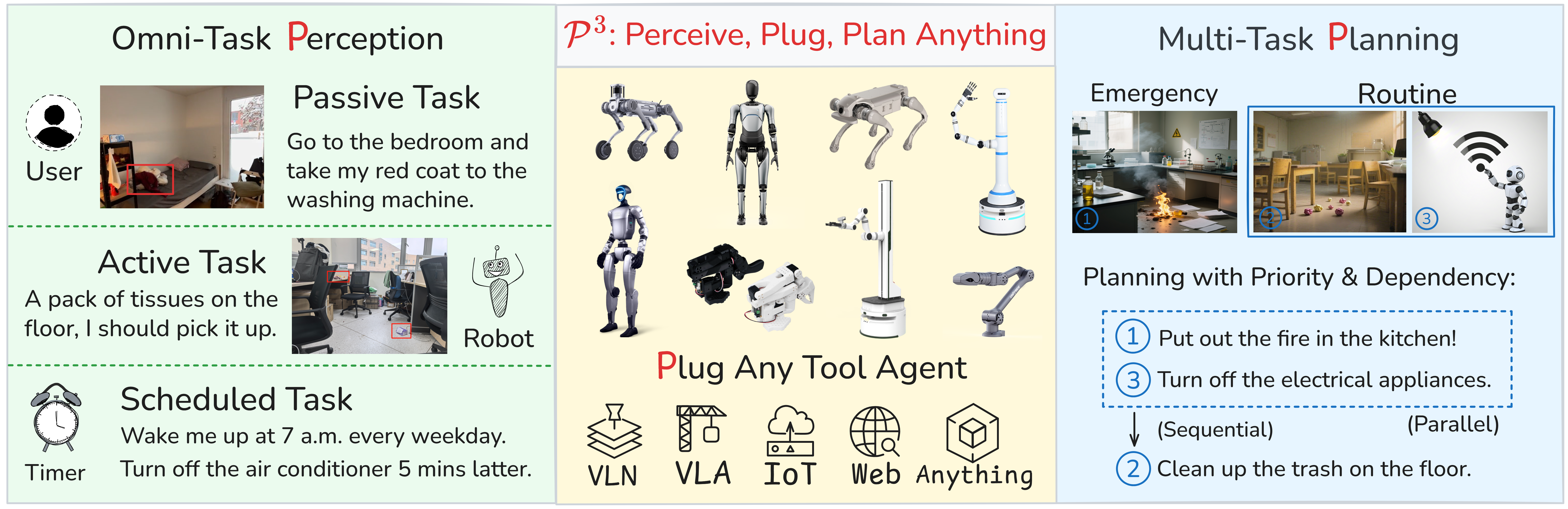}
    \caption{Overview of ${\mathcal{P}}^3$. The framework centers on three capabilities: \textbf{Perceive}, where the Brain Agent interprets streamed environments through the Perception Module; \textbf{Plug}, where IoT, web, navigation, and manipulation tools can be integrated without requiring feedback; and \textbf{Plan}, where the Task Planner manages multi-task execution based on memory, priorities, dependencies, and user instructions. Purple highlights denote the proposed modules.}
    \label{fig:teaser}
\end{figure*}

Embodied agents are intelligent systems that perceive, reason, and act through a physical or virtual body to interact with their environment in real time~\cite{liu2024aligningcyberspacephysical}. The development of such agents aims to bridge perception, reasoning, and action, enabling seamless interaction with the environment.
Prior research in embodied intelligence and multi-modal perception has achieved promising results across various tasks~\cite{lin2024bip3d, roboos, tan2024cradleempoweringfoundationagents}.
However, these works rely solely on feedback from tools, resulting in incomplete environment observations, restrictions on tool usage, and limited task settings. These limitations create a significant gap between agent capabilities and real-world task demands.
To address this challenge and develop versatile embodied agents, a more generalized architecture is needed.

Specifically, a key limitation of prior embodied agents~\cite{yang2025agenticrobotbraininspiredframework, roboos, wang2024e2clexplorationbasederrorcorrection,fan2025embodiedvideoagentpersistentmemory} stems from their passive environmental perception mechanism.
For example, during a cleaning task, a robot may fail to adapt if a person suddenly moves furniture, disrupting its plan.
As these systems mistakenly assume that only their own actions alter the environment, they rely on task feedback instead of detecting real-time changes caused by external factors, such as human or other agent actions.
Moreover, previous embodied agents presuppose reliable feedback from tool agents (i.e., external modules or APIs that provide specific functionalities), which in turn limits the set of available operations to predefined toolkits. 
In practice, many real-world edge devices fail to provide such feedback for several reasons. First, a large class of devices, such as Internet of Things (IoT) components and basic mechanical devices like motors, are inherently designed without bidirectional communication~\cite{aouedi2024surveyintelligentinternetthings}. Second, certain IoT devices intentionally restrict themselves to one-way communication as a security measure~\cite{securing_iot_edge, cybersecurity_challenges}. Finally, resource constraints on edge hardware make it impractical to host complex communication stacks or protocol negotiation~\cite{unidirectional_communications}. Collectively, these compatibility constraints materially hinder the practical deployment of tool-dependent embodied agents.

To address these issues, we introduce a perception module that actively detects dynamic changes in the environment, including those caused by the agent's own actions, human behaviors, other agents' activities, and dynamic alterations in context. 
In contrast to previous systems with fragmented perception~\cite{Being-0, Nazarczuk2024ClosedLI}, our design decouples the perception module from specific tool agents and treats it as an always-on centralized service that continuously observes and perceives environmental cues, thereby enabling effective deployment of embodied agents in real-world environments. These cues are stored and periodically fed into the Visual Language Models (VLMs)~\cite{Qwen2.5-VL}, where context engineering is applied to retrieve dynamic events in the scene, identify tasks actively, and update the scene graph for understanding high-level spatial relations.
\textbf{This architecture brings the following benefits}: (1) agents can detect changes that tool-based systems would miss, enhancing their ability to respond to unexpected events; (2) agents can actively propose tasks based on their continuous observations, such as room cleaning or emergency handling; and (3) agents can seamlessly integrate with any tool, even those lacking standard feedback mechanisms.

Owing to the perception module, a wide range of active tasks (i.e., tasks that are proactively proposed by the perception module) and diverse tool plug-ins are introduced, boosting the range and complexity of executable tasks and available tools.
This broader scope further exposes the weaknesses of earlier task-planning methods.
Specifically, earlier task planning methods typically follow a fixed first-in-first-out manner, lacking the ability to manage task executions based on user preferences or task dependency.
For example, such systems would insist on finishing room tidying before addressing urgent events like putting out a fire.

To overcome these shortcomings and build versatile embodied agents, it is crucial to plan and manage omni-tasks (i.e., active, passive, and scheduled tasks, as shown in Fig.~\ref{fig:teaser}) by considering dependencies, user priorities, and environmental context. Such agents must also support flexible task execution, interruption, and resumption.
In response, we propose a novel task planning module that jointly processes all types of tasks. This module evaluates task priorities and dependencies, while also incorporating a task memory to store pending, scheduled, and interrupted tasks. The planning module selects the available tasks with the highest priority for execution, enabling our \textbf{Perceive–Plug–Plan (${\mathcal P}^3$) framework} to effectively handle complex real-world task scheduling.

In summary, our main contributions are:

\begin{itemize}
    \item \textbf{Omni-task Perception}: We introduce a unified perception module that actively detects real-time environmental changes, allowing the agent to autonomously identify various types of tasks based on dynamic context and environmental shifts without passive feedback.
    \item \textbf{Plug-any-tool Agent}: By removing the need for tool feedback, we enable seamless integration with a wide range of devices, including those without standardized interfaces, enhancing the agent's flexibility and compatibility with diverse real-world tools.
    \item \textbf{Multi-task Planning}: We propose a dynamic task planner that effectively manages multiple tasks, autonomously arranging them based on real-time context and dependencies, ensuring flexible task execution in complex environments.
    \item \textbf{Real-world deployments}: We evaluate the strong performance of our ${\mathcal P}^3$ framework by testing it through real-world deployments, demonstrating its ability to handle dynamic, real-world challenges effectively.
\end{itemize}

\section{Related Work}

\begin{figure*}[tbp]
    \centering
    \includegraphics[width=0.96\textwidth]{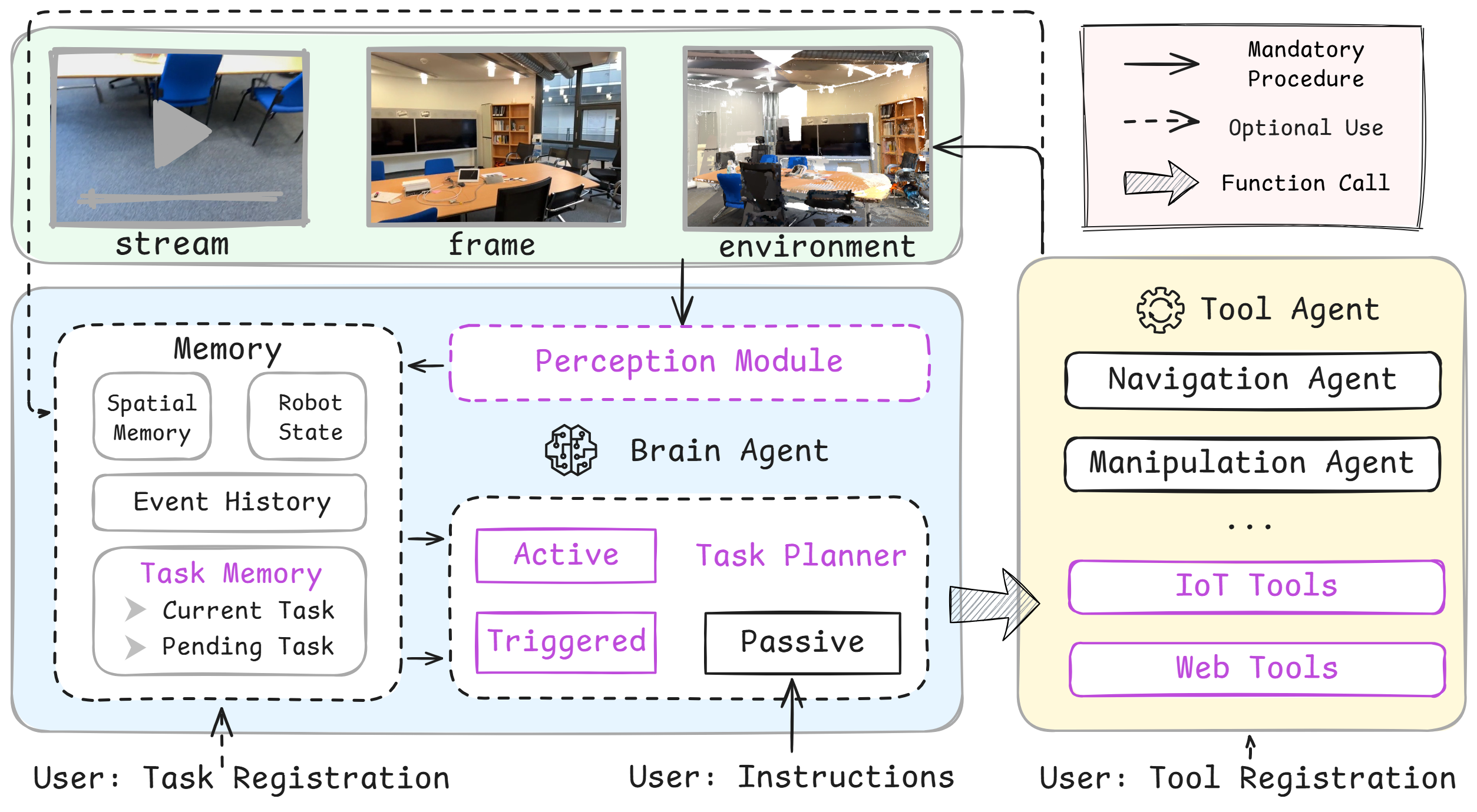}
    \caption{Overall architecture of ${\mathcal{P}}^3$. The framework combines active perception, open-ended tool plug-ins, and LLM-guided multi-task planning. Purple highlights denote the proposed modules.}
    \label{fig:framework}
\end{figure*}

\noindent \textbf{Embodied System.}
Recent advances in embodied intelligence have focused on developing integrated, modular, and scalable robotic systems. Early frameworks like the Robot Operating System (ROS)~\cite{ros} introduced modularity with a node-based architecture, facilitating the coordination of complex behaviors. Building on this foundation, next-generation platforms such as RoboOS~\cite{roboos} have adopted a hierarchical ``Brain-Cerebellum'' paradigm, which merges large vision-language models for high-level reasoning with a modular skill library for low-level execution. Likewise, frameworks like RAI~\cite{rai} and AllenAct~\cite{allenact} emphasize interoperability with both simulated and physical environments, providing flexible infrastructures for learning-based embodied tasks. Concurrently, generalist agents such as Being-0~\cite{Being-0}, Gemini Robotics~\cite{geminirobotics}, and $\pi_0$~\cite{pi0} strive to integrate vision, language, and action, demonstrating strong generalization across various physical tasks without the need for explicit retraining.
Despite recent advances, many systems still struggle with limited perception and poor tool compatibility. They often overlook external factors, relying solely on task feedback to infer scene changes, thus resulting in partial observations and accumulated errors. Moreover, their dependence on fixed hardware and skill sets hinders adaptation to novel tools or unstructured tasks. In contrast, we propose a unified perception module and a plug-any-tool architecture that actively detects scene dynamics and supports seamless integration of diverse tools and tasks, significantly enhancing the agent’s adaptability and scalability in real-world deployments.

\noindent \textbf{Embodied Tasks.}
Most existing embodied tasks focus on passive task execution, where agents follow explicit human instructions rather than acting autonomously. Paradigms like Vision-Language Navigation~\cite{vlnr1,lovon} and Vision-Language Action~\cite{unifiedvla,lbmtri2025,xu2025a0affordanceawarehierarchicalmodel} reflect this trend, focusing on instruction-following in structured settings with predefined goals. 
However, by centering on user-provided commands, these frameworks offer no mechanisms to generate or pursue tasks autonomously, limiting their adaptability in dynamic environments.
In contrast, active tasks, such as emergency response~\cite{triffid} or room cleaning~\cite{kant2022housekeep}, require agents to perceive their surroundings, identify task opportunities, and act without explicit guidance. Nevertheless, most existing systems~\cite{tidybot, li2024llmenhancedscenegraphlearning} for active tasks are constrained by limited scene scopes and task range, which oversimplify the complexities of real-world interactions. Our work addresses these limitations by introducing a robust framework that supports both passive and active tasks.
By integrating a real-time task planner, we make agents more capable of handling various types of tasks in open environments.
\section{Method} \label{sec:method}

In this section, we introduce the $\mathcal P^3$ framework for versatile embodied agents, covering real-time omni-task \textbf{perception}, open-ended tool \textbf{plug-in}, and multi-task \textbf{planning}. The overall framework is illustrated in Fig.~\ref{fig:framework}.
 
To remain aware of changes caused by both the agent's own actions and external factors, we decouple perception from individual tool agents and integrate it into a centralized Perception Module.
The module leverages vision-language models to proactively observe scene dynamics, detect events, propose active tasks, and update the scene graph for a high-level spatial representation. By eliminating the reliance on tool agents’ feedback, the system gains compatibility with any real-world devices, especially those lacking feedback.

\begin{figure*}[htbp]
    \centering
    \includegraphics[width=0.94\textwidth,keepaspectratio]{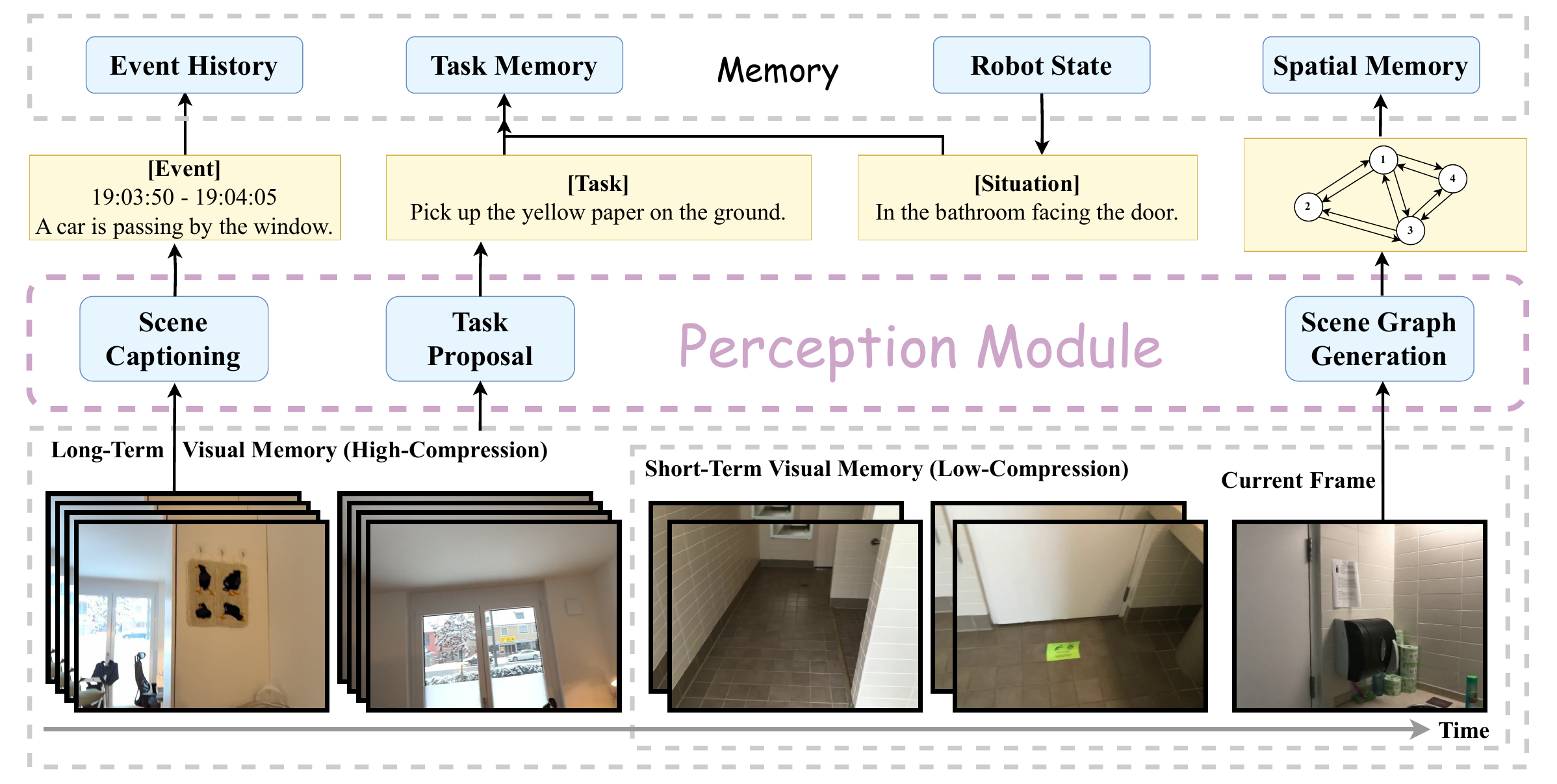}
    \caption{Internal structure of the Perception Module in $\mathcal P^3$. It processes visual streams across time using both long-term (high-compression) and short-term (low-compression) visual memories. Three key functions, Scene Captioning, Task Proposal, and Scene Graph Generation, convert visual input into symbolic memory forms, supporting perception-driven task planning.}
    \label{fig:perception}
\end{figure*}

Meanwhile, we adopt an LLM for task planning, as large-scale language pretraining empowers it to capture task importance, dependency patterns, and stronger logical reasoning.
The planner collaboratively organizes active tasks proposed by the perception module, triggered by a timer from the task memory, and passive tasks from users' instructions.
These tasks are jointly evaluated according to priority and dependency, allowing the planner to select which tasks should be executed immediately. Acting like an air-traffic controller, the planner arranges the selected tasks and hands them off to a dispatcher, assigning each job to the appropriate tool agent for execution.
Finally, during execution, although tool agents no longer provide feedback to agents, their impact on the environment is captured by the perception module, which further aids the decision-making process of the next step.

\subsection{Active Perception for Omni-Task Awareness}
Similar to humans’ continuous visual perception, $\mathcal P^3$ employs an active perception module that decouples sensing from tool agents and serves as a centralized, always-on observer. This module leverages vision–language models to continuously analyze streamed video input, detect scene changes, infer causal events, and generate context-aware task proposals. 

Each new frame updates the scene graph that encodes spatial relations among objects and agents~\cite{wang2024allseeing_v2}. Temporal reasoning is then performed by the VLM through prompt-based context engineering to identify event patterns—such as “a door left open” or “a person entering the room.” These detected events are translated into structured task proposals (e.g., “close the door,” “greet the visitor”), which are stored in the task memory for future scheduling.

To further mitigate context-window limitations, we employ hierarchical memory buffers and on-the-fly summarization: the current query and previous urgent events will be fully sent to the planner, while long-term patterns are archived in an external knowledge store and retrieved. This guarantees that only the most relevant information occupies the LLM’s prompt, preserving low latency and robust performance.
As illustrated in Fig.~\ref{fig:perception}, this architecture enables $\mathcal P^3$ to proactively capture environmental dynamics, generate context-aware active task proposals, and maintain event histories that support omni-task perception.

\subsection{Open-Ended Tool Plug-ins}

The second core innovation of $\mathcal P^3$ is its plug-any-tool mechanism, which breaks the dependency on bidirectional feedback. In previous embodied agents, navigation (VLN) and manipulation (VLA) models served as foundational tools for task execution. In $\mathcal P^3$, they remain essential agents as they directly control the robot’s movements and actions. For safety and reliability, all commands sent to the manipulation agent undergo validation prior to execution. These checks are not saved to memory as the isolation mechanism prevents conflicts and interference between information and helps control the context length.
Apart from these mandatory agents, our framework further supports a wide range of real-world tools, including a wide range of IoT devices and web agents, without relying on bidirectional feedback loops. By leveraging context engineering techniques (e.g., dynamic scene summarization and relevance filtering), $\mathcal P^3$ maintains a compact yet informative record of past observations. This allows it to issue commands like turning off lights or adjusting air-conditioning without physically moving when the user leaves for work. Similarly, by incorporating timely updates from web agents, it can advise the user to bring an umbrella when rain is forecast.
Moreover, the perception module eliminates the need for feedback from tool agents, enabling users to register customized tools directly. A tool agent can be added simply by specifying its function and corresponding function-call interface. This design ensures that any controllable element in the environment can be seamlessly incorporated into $\mathcal P^3$.

\subsection{LLM-Guided Multi-Task Planning}

To address the shortcomings of first-in-first-out scheduling in previous embodied robotics systems, we propose a novel task planning module that jointly considers task importance and inter-task dependencies. Specifically, we categorize tasks into three types: passive, active, and scheduled. Passive tasks (e.g., ``open the window'') are triggered by users' instructions or queries, while Active tasks (e.g., ``tidy up the room'') are proposed by the perception module. Scheduled tasks (e.g., ``turn on the air-conditioner at 5 p.m'' and ``monitor the patient's temperature hourly'') are registered by the user and stored in the task memory. Scheduled tasks are triggered by the timer according to users' instructions. 
When any task is triggered, all tasks that can be executed instantly (i.e., not pending for the timer) are sent into an LLM for task priority and dependency evaluation.
Moreover, by retrieving relevant events and prior states from memory, the scheduler obtains contextual cues that improve judgment. Finally, available high-priority tasks are dispatched through the dispatcher.

\subsection{Memory Modules}

Our memory design further supports the framework by making high-level language instructions executable in dynamic, multi-room environments. It couples a semantically labeled spatial map with a symbolic scene representation and temporally indexed traces so the planner can ground references like ``go to the desk in the office'', monitor scene changes, and resume interrupted work. This design follows our Memory block that interfaces with the perception module’s scene graph and two visual buffers, as illustrated in Fig.~\ref{fig:perception} and discussed in Sec.~\ref{sec:method}.

\textbf{Spatial Memory.} Spatial Memory stores the scene graph. We initialize this graph with a manual scan-and-annotation pass that assigns room names, such as laboratory and office, to prevent linguistic ambiguity during grounding. This setup matches our real deployments, which span a research laboratory and a typical office, and it ensures that instructions like ``go to the chair in the lab'' resolve against named regions rather than ambiguous geometry.
At run time, every incoming frame is routed to the Perception Module for relation parsing in the specified area. The module applies ASMv2~\cite{wang2024allseeing_v2} to infer object-to-region and object-to-object spatial relations, then commits those relations as edges in the scene graph that lives inside Spatial Memory.

\textbf{Event History.} Event History records a time-stamped stream of changes, such as a door being left open, an object being knocked over, or a person entering. This history is the evidence for causal reasoning and priority updates, for example, elevating a safety-relevant cleanup that appears mid-route. It also prevents redundant proposals by recognizing that a condition was already handled. This mechanism reflects the detection of events and the storage of event histories, which are critical when active or scheduled tasks arrive during execution.

\textbf{Task Memory.} Task Memory is the planner's list of current, pending, and scheduled tasks that enables interruption and resumption with full context. When a higher-priority instruction arrives, the current job is snapshot with its targets and partial progress, then later reinstated at the same step. This is essential for competing task scenarios where user instructions, perception of proposed jobs, and timers interact. The experiments highlight that a dynamic scheduler working with task memory supports interruption and resumption in multi-room routines.

\textbf{Robot State.} Robot State stores instantaneous capabilities and constraints, including pose, gripper occupancy, battery, and safety interlocks. Keeping this in memory makes feasibility visible to the planner and reduces futile dispatches, for example, avoiding a grasp when the gripper is occupied. Robot State appears as part of the Memory block and is consumed by the planner during scheduling.

By integrating these complementary memories into a single operational loop, our $\mathcal P^3$ framework achieves precise language grounding, temporal consistency, and physically feasible execution in real environments. The scene graph in Spatial Memory removes ambiguity at the room and object level, event history supplies the temporal context for causal updates and priority reordering, task memory preserves progress for interruption and resumption, and robot state enforces safety and feasibility during dispatch. Together, these components allow $\mathcal P^3$ to plan over long horizons, recover from disturbances, and generalize across hardware platforms and layouts in complex, multi-room settings. In practice, this yields reliable completion of passive, active, and scheduled tasks under real-world noise and partial observability, including navigation to named regions, object search, and the seamless handoff between passive user commands and active routines.
\section{Experiments}

\subsection{Experimental Settings}

In this section, we evaluate whether $\mathcal P^3$ delivers the three capabilities targeted by the framework: active perception, plug-any-tool execution, and competing-task planning. We deploy the framework on humanoid dual-arm robots from Realman and AgiBot and test it in real-world task settings.

For manipulation, we integrate the Moka pipeline~\cite{fangandliu2024moka}, a high-precision grasping solution, to assess object grasping and precise physical execution. For navigation, we use the Woosh chassis~\cite{woosh_samir50}, which follows a SLAM-based approach for multi-step paths in dynamic environments. We also connect Xiaomi Smart Home devices to evaluate IoT interaction and demonstrate compatibility with common real-world appliances.

\begin{table*}[ht]
\centering
\caption{Standard Configuration of the RealMan Dual-arm Lift Platform}
\label{tab:realman_config}

\begin{tabular}{@{}l@{\hspace{0.8cm}}c@{\hspace{0.8cm}}p{5cm}@{}}
\toprule
\textbf{Item} & \textbf{Quantity} & \textbf{Specification} \\
\midrule
\multicolumn{3}{l}{\textit{Main Control}} \\
\quad Jetson AGX Orin & 1 & 64 GB RAM + 1 TB SSD \\
\midrule
\multicolumn{3}{l}{\textit{Manipulators}} \\
\quad RM65-B-V robotic arm & 2 & 6-DOF \\
\quad RMG24 two-finger parallel gripper & 2 & stroke 24 mm \\
\midrule
\multicolumn{3}{l}{\textit{Vision System}} \\
\quad Intel RealSense D435 stereo camera & 3 & depth + RGB \\
\quad RGB surveillance camera & 2 & fixed installation \\
\midrule
\multicolumn{3}{l}{\textit{Mobility \& Power}} \\
\quad Mobile chassis & 1 & differential drive \\
\quad LiFePO$_4$ battery & 1 & 48 V, 20 Ah \\
\quad Electric lifting column & 1 & vertical travel 0–2200 mm \\
\midrule
\multicolumn{3}{l}{\textit{Dimensions}} \\
\quad Overall size (folded) & & 640 × 450 × 1700 mm \\
\quad Weight & &  90 kg \\
\quad Working range & & vertical 0–2200 mm, horizontal 0–1900 mm, forward 0–600 mm \\
\bottomrule
\end{tabular}
\end{table*}
\begin{table*}[!ht]
\centering
\caption{Detailed Action Space of $\bm{\mathcal{P}}^3$ Functional Modules}
\label{tab:action_space_roboagent}
\small
\renewcommand{\arraystretch}{1.3}
\begin{tabularx}{\textwidth}{|c|c|X|X|X|}
\hline
\textbf{Module} & \textbf{Agent} & \textbf{Functionality} & \textbf{Input} & \textbf{Output} \\
\hline

\multirow{2}{*}{Perception} 
& \multirow{1}{*}{Perception Module} 
  & Environment Recognition & \(\mathrm{camera}_{id}\) & Image stream \\

\cline{2-5}
 & \multirow{1}{*}{MicrophoneAgent} 
& Live Listening & -- & Continuous microphone stream \\

\hline
\multirow{7}{*}{Interaction} 
& \multirow{3}{*}{SpeakerAgent} 
  & Text-to-Speech & Response text & Generated voice audio \\
& & Multilingual Output & Text + target language & Spoken in selected language \\
& & Emotional Expression & Text + emotion label & Expressive voice synthesis \\

\cline{2-5}
& \multirow{4}{*}{WebAgent} 
  & Weather Query & Location / city name & Weather report \\
& & News Retrieval & Topic / category (optional) & News headlines and summaries \\
& & Parcel Service & Shipment request / tracking ID & Delivery status or confirmation \\
& & Time Query & Timezone / location (optional) & Current time \\
  
\hline
\multirow{10}{*}{Control} 
& \multirow{5}{*}{NaviAgent} 
  & Target Navigation & Destination or coordinates & -- \\
& & Local Movement & Direction + distance  & Robot movement \\
& & Local Rotation & Angle + direction & Robot orientation update \\
& & Navigation Interruption & -- & Halt and task switch \\
& & Position Query & -- & Current position \\
  
\cline{2-5}
& \multirow{2}{*}{ManiAgent} 
  & Object Grasping & Object description & Grasp success / failure \\
& & Operation Interruption & Emergency / high-priority task & Mid-task stop \\
  
\cline{2-5}
& \multirow{3}{*}{IoTAgent} 
  & Light Control & command and brightness level & -- \\
& & Humidifier Control & command and humidity level & -- \\
& & Device State Query & Device name & Current status \\
\hline

\end{tabularx}
\end{table*}

For the real-world deployment studies, we report the robot configuration, tool library, assigned task pool, task-level execution details, key hyperparameters, and implementation settings to make the experimental setup reproducible. To maintain a fair and reliable comparison across tasks, all physical experiments reported in Table~\ref{tab:real-world} were conducted using the RealMan dual-arm lifting platform. Its complete hardware and software configuration is listed in Table \ref{tab:realman_config}, including sensors, computation modules, and gripper settings. To further verify the generality and transferability of our framework, we also deployed the same system on the Agibot platform.

\begin{table*}[htbp]
\centering
\caption{Assigned Proactive and Scheduled Tasks}
\label{tab:assign_task}
\begin{tabular}{lp{3cm}p{7.5cm}p{4.5cm}}
    \toprule
        No. & Task & Description & Time  \\
    \midrule
    \rowcolor{gray!20} \multicolumn{4}{l}{\textbf{Active Tasks}} \\
    \midrule
    1 & Debris Removal    & Cleans up loose fragments without a set place.                      & Anytime.  \\ 
    2 & Item Organization & Organizes misplaced items into their spots.              & Anytime.  \\ 
    3 & Safety Check      & Quickly inspects for potential hazards.                             & Anytime.  \\ 
    \midrule
    \rowcolor{gray!20} \multicolumn{4}{l}{\textbf{Scheduled Tasks}} \\
    \midrule
    4 & Go to the office sofa  & Uses VLN to move to the office sofa.        & Every Friday at 5:30 PM.  \\ 
    5 & Exercise reminder & Triggers through the speaker to remind the user to exercise.   & Every weekday at 9 PM.    \\ 
    6 & Weather broadcast     & Fetches weather data via web agent and announces it. & Every morning at 9 AM.    \\ 
    7 & Watering reminder     & Sends a reminder to water plants. & Every Sunday at 10 AM.   \\ 
8 & Medicine reminder  & Reminds user to take medicine via text alert. & Everyday at 7 AM. \\ 
9 & Meeting reminder & Notifies user of the weekly meeting schedule. & Every Monday at 9:30 AM. \\ 
    \bottomrule
\end{tabular}
\end{table*}

\begin{figure*}[htbp]
    \centering
    \includegraphics[width=0.93\textwidth]{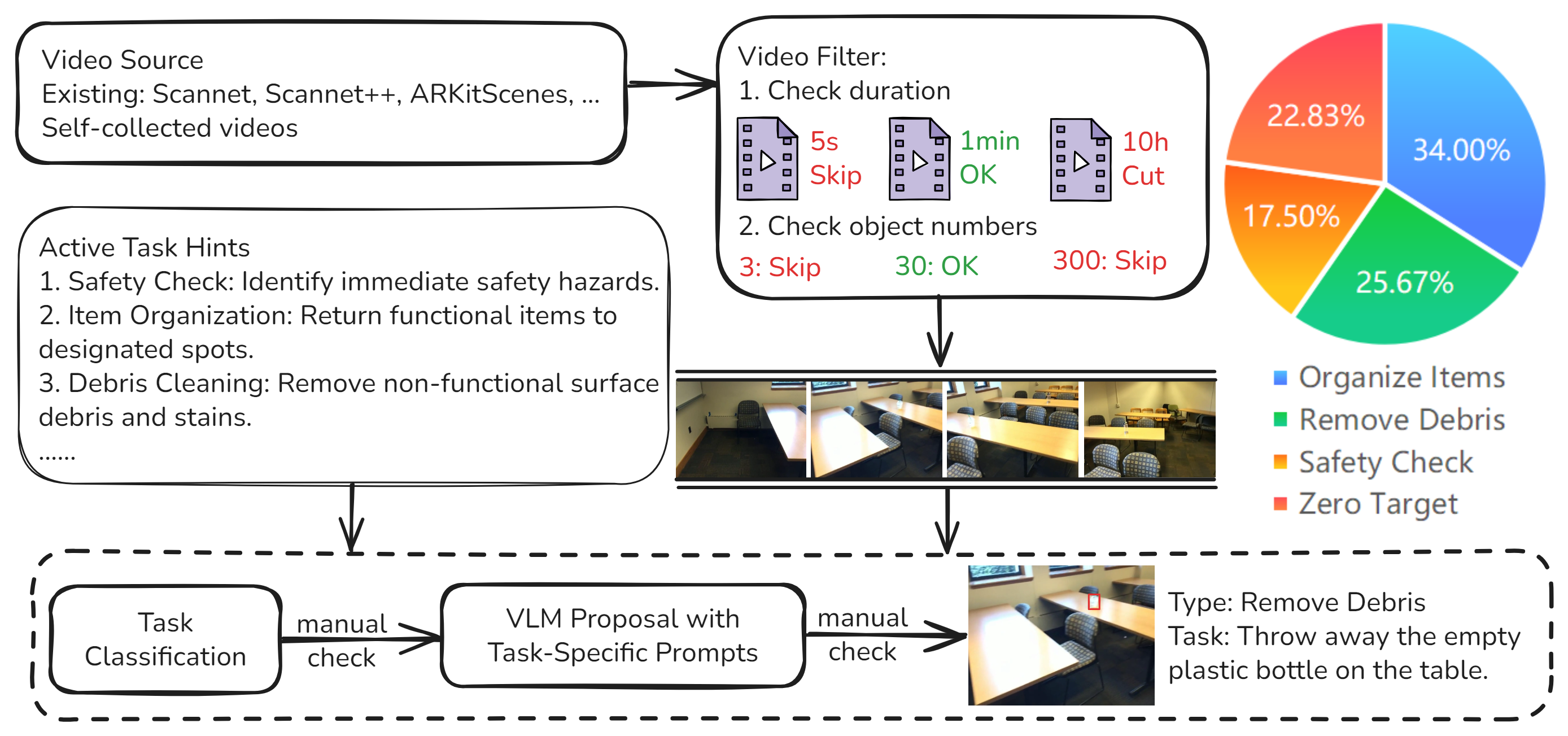}
    \caption{The pipeline for collecting and annotating data for the Active Task Perception Benchmark.}
    \label{fig:benchmark}
\end{figure*}

Prior to the experiments, we equipped the robot with a versatile tool set so that it could handle a wide range of tasks. The full action and skill library available to the system is summarized in Table \ref{tab:action_space_roboagent}, covering both atomic and composite actions. All trials take place in two realistic indoor environments: a research laboratory and a typical office. To simulate diverse use cases, the robot is not only given reactive commands triggered by user input but also assigned proactive and scheduled tasks that require planning and contextual reasoning, which are summarized in Table \ref{tab:assign_task}. In Table \ref{tab:independent} and \ref{tab:compating}, we provide a detailed breakdown of the eleven evaluated tasks, including task descriptions, initial conditions, tool requirements, and success criteria.

The following experiments are organized around these three capabilities: active perception, tool plug-in and execution, and competing-task planning under priorities and dependencies.

\subsection{Active Perception Capability Evaluation}

As the $\mathcal P^3$ framework requires the perception module to observe and propose potential tasks in the environment, it is vital that the VLMs used for the perception module can meet the requirement. To evaluate such a capability, we built the Active Task Perception Benchmark. First, we collect videos from publicly available datasets~\cite{scannet,scannet++,yang2024think,dehghan2021arkitscenes}, while also shooting original videos with proper duration.
Then, combined with active task hints (e.g., safety check, item organization), the videos undergo task classification, followed by generating VLM proposals with task-specific prompts. Both steps involve manual checks. Finally, we construct a benchmark that consists of 600 snapshots from these processed first-person videos. These snapshots are preprocessed by GPT-4o~\cite{gpt4o} and QwenVL~\cite{Qwen2.5-VL} to generate task proposals and bounding boxes for reference, and are then manually checked again to ensure data quality. 

In our benchmark, we evaluate three representative categories of household tasks: debris cleaning, item organization, and safety checks.  
To prevent the situation where the model achieves high performance by arbitrarily guessing tasks, we add negative samples where the room is tidy (i.e., no task should be proposed).
As the perception module should be adaptable to all open-ended tasks, we utilize several off-the-shelf VLMs trained on either general datasets or embodied tasks and evaluate them under a zero-shot setting.
Finally, we measure accuracy using \texttt{ChatGPT-4.1-0414}~\cite{gpt3.5} for judgment, together with the average number of proposals on negative samples.

To construct the Active Task Perception Benchmark, we design a collection and annotation pipeline that combines public egocentric videos, original recordings, VLM-assisted proposal generation, and manual verification, as summarized in Fig.~\ref{fig:benchmark}.

\begin{table*}[!ht]
    \centering
    \caption{Snapshot Examples in the Benchmark}
    \label{tab:benchmark_demo}
    \vskip 0.12in
    \hrule
    \vskip 0.10in
    \begin{minipage}{0.24\textwidth}
        \centering
        \textbf{Debris Cleaning}
    \end{minipage}
    \begin{minipage}{0.24\textwidth}
        \centering
        \textbf{Item Organization}
    \end{minipage}
    \begin{minipage}{0.24\textwidth}
        \centering
        \textbf{Safety Check}
    \end{minipage}
    \begin{minipage}{0.24\textwidth}
        \centering
        \textbf{Negative Sample}
    \end{minipage}\hfill
    
    \vskip 0.10in
    \hrule
    \vskip 0.10in
    
    \begin{minipage}{0.24\textwidth}
        \centering
        \includegraphics[width=0.9\linewidth]{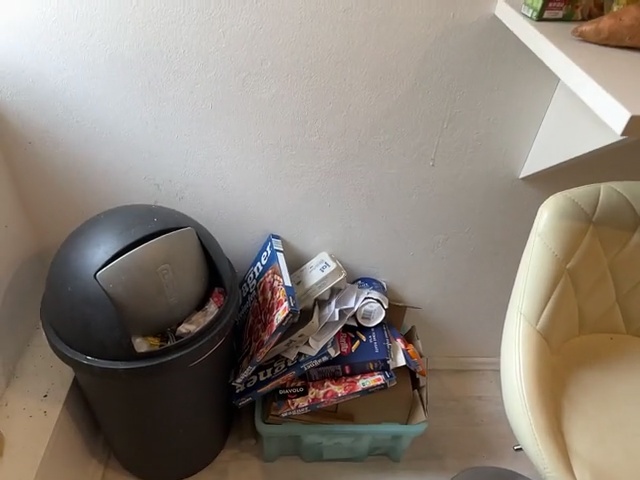}
    \end{minipage}
    \begin{minipage}{0.24\textwidth}
        \centering
        \includegraphics[width=0.9\linewidth]{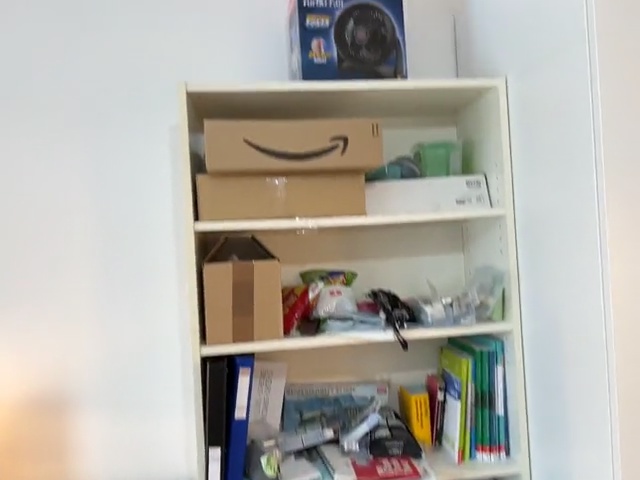}
    \end{minipage}
    \begin{minipage}{0.24\textwidth}
        \centering
        \includegraphics[width=0.9\linewidth]{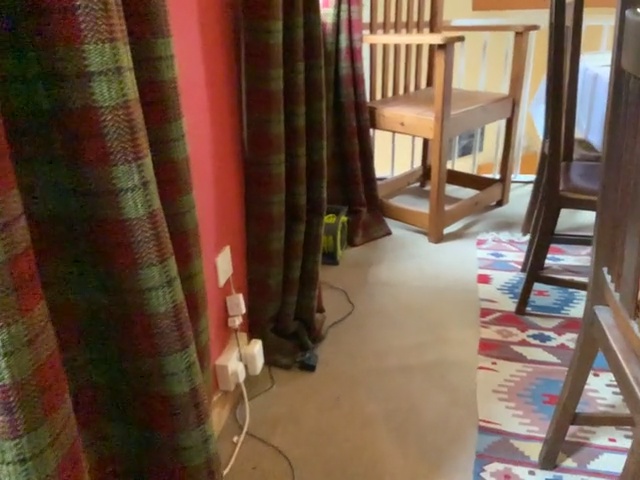}
    \end{minipage}
    \begin{minipage}{0.24\textwidth}
        \centering
        \includegraphics[width=0.9\linewidth]{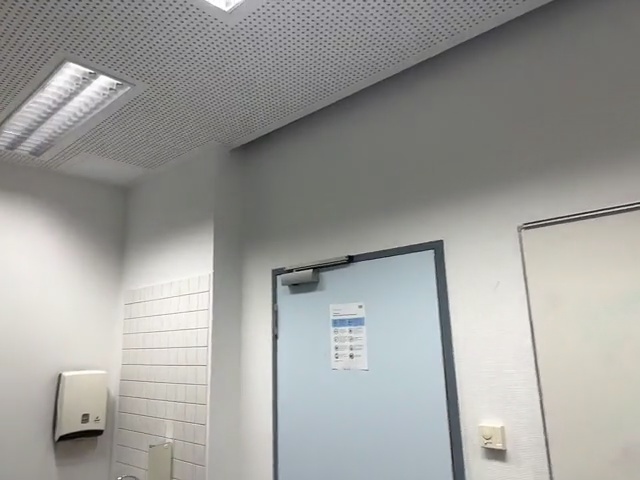}
    \end{minipage}\hfill
    
    \vskip 0.10in
    \hrule
    \vskip 0.10in
    
    \begin{minipage}{0.24\textwidth}
        \centering
        \includegraphics[width=0.9\linewidth]{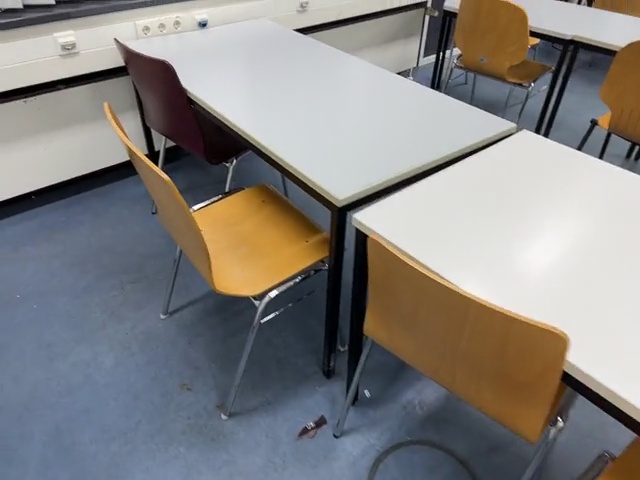}
    \end{minipage}
    \begin{minipage}{0.24\textwidth}
        \centering
        \includegraphics[width=0.9\linewidth]{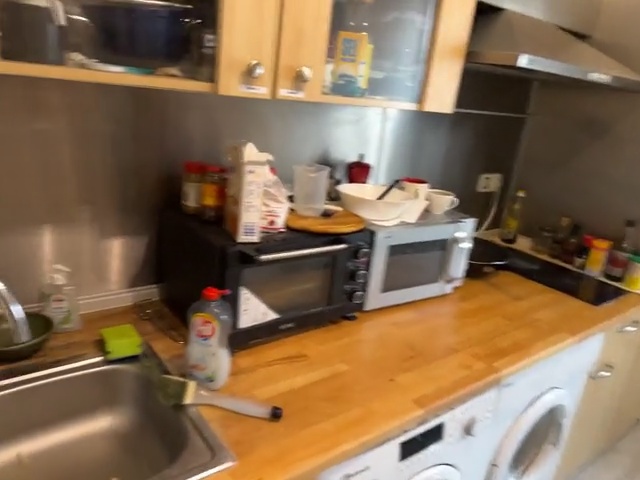}
    \end{minipage}
    \begin{minipage}{0.24\textwidth}
        \centering
        \includegraphics[width=0.9\linewidth]{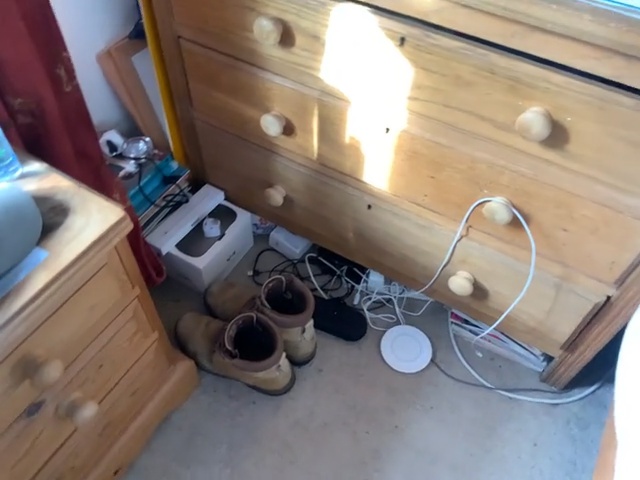}
    \end{minipage}
    \begin{minipage}{0.24\textwidth}
        \centering
        \includegraphics[width=0.9\linewidth]{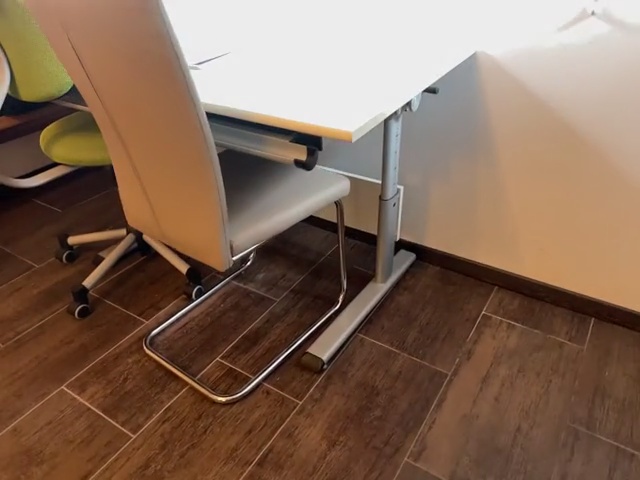}
    \end{minipage}\hfill
    
    \vskip 0.10in
    \hrule
    \vskip 0.10in
    
    \begin{minipage}{0.24\textwidth}
        \centering
        \includegraphics[width=0.9\linewidth]{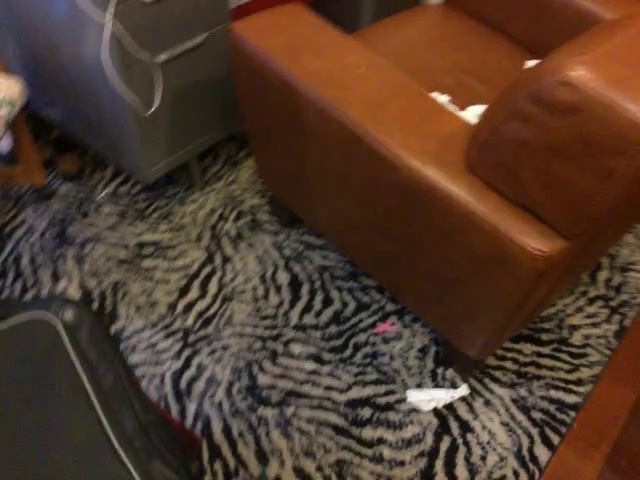}
    \end{minipage}
    \begin{minipage}{0.24\textwidth}
        \centering
        \includegraphics[width=0.9\linewidth]{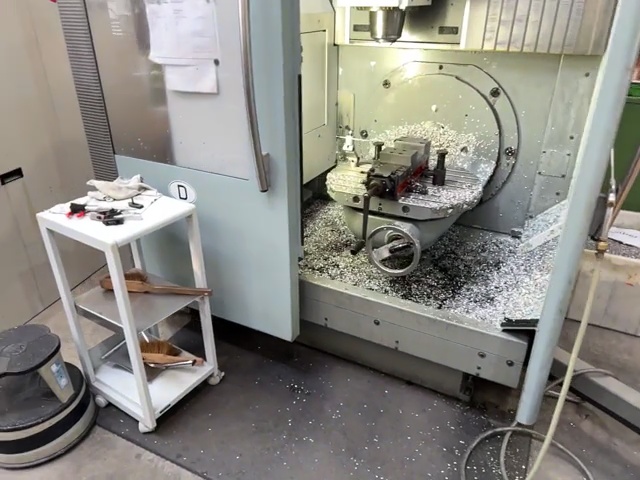}
    \end{minipage}
    \begin{minipage}{0.24\textwidth}
        \centering
        \includegraphics[width=0.9\linewidth]{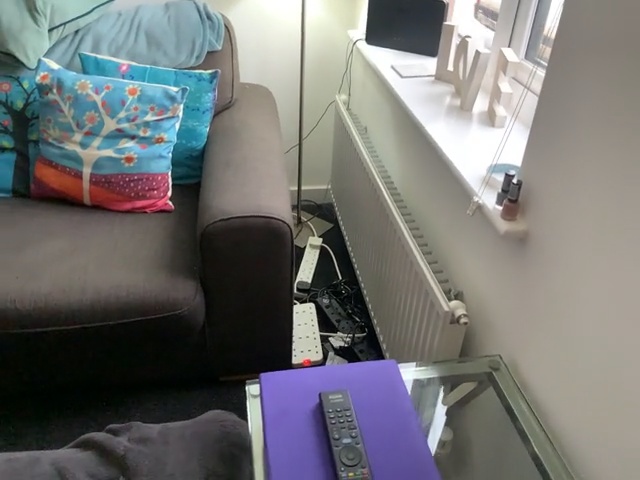}
    \end{minipage}
    \begin{minipage}{0.24\textwidth}
        \centering
        \includegraphics[width=0.9\linewidth]{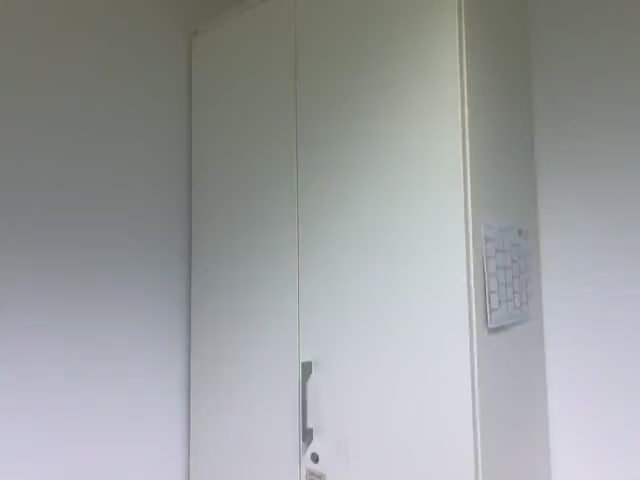}
    \end{minipage}\hfill
    
    \vskip 0.10in
    \hrule
    \vskip 0.10in
    
    \begin{minipage}{0.24\textwidth}
        \centering
        \includegraphics[width=0.9\linewidth]{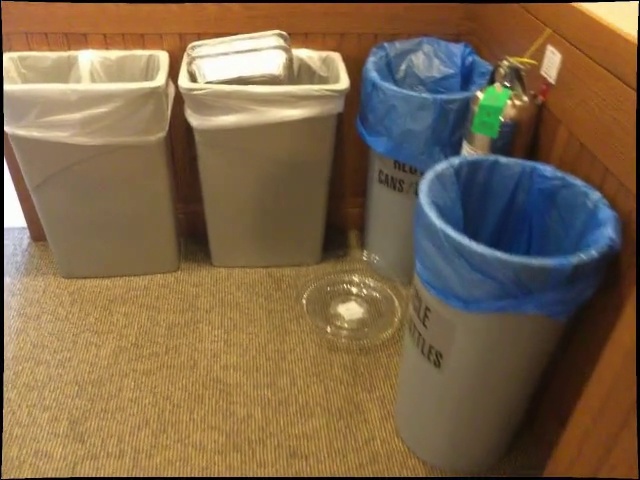}
    \end{minipage}
    \begin{minipage}{0.24\textwidth}
        \centering
        \includegraphics[width=0.9\linewidth]{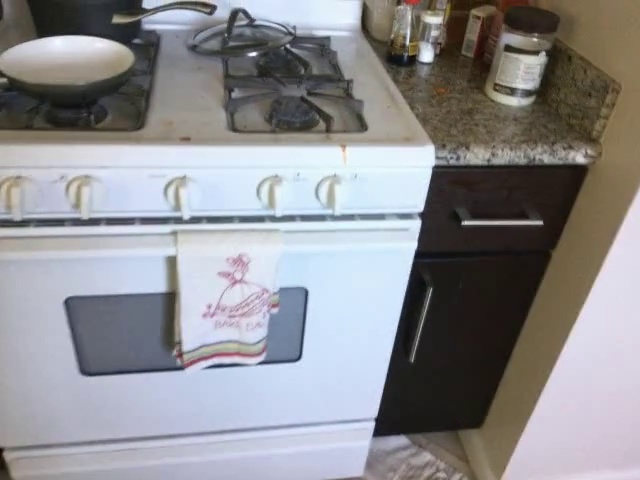}
    \end{minipage}
    \begin{minipage}{0.24\textwidth}
        \centering
        \includegraphics[width=0.9\linewidth]{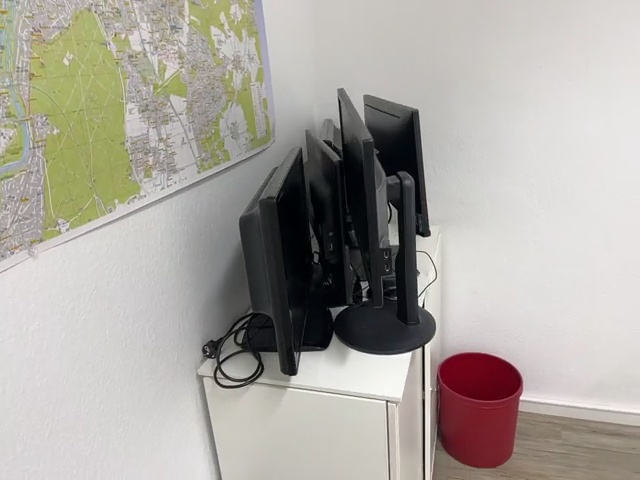}
    \end{minipage}
    \begin{minipage}{0.24\textwidth}
        \centering
        \includegraphics[width=0.9\linewidth]{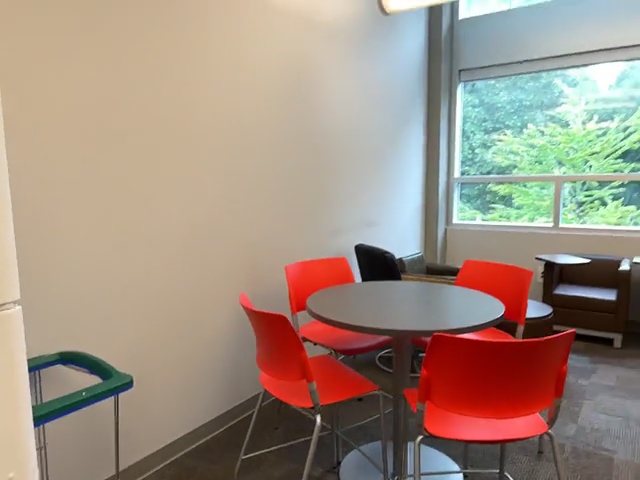}
    \end{minipage}\hfill
    
    \vskip 0.10in
    \hrule
    
\end{table*}

The data collection stage starts from a series of first-person videos from publicly available datasets~\cite{scannet,scannet++,yang2024think,dehghan2021arkitscenes}. Additionally, to further enhance the diversity of our benchmark, we also record videos under the same basic setting (first-person, similar length, etc.).
To ensure that the videos can reflect real-world scenarios, we filter the videos and preserve those with appropriate lengths and numbers of objects.
Then, for each video, we take a snapshot every 3 seconds.
By applying state-of-the-art VLMs~\cite{Qwen2.5-VL, gpt4o} and manual check, we select a subset of 600 snapshots belonging to the following categories:

\begin{itemize}
    \item \textbf{Debris Cleaning}: Removes all non-functional surface contaminants—both loose solid fragments and adhered liquids or stains. Simple removal or wiping restores cleanliness, excluding functional items or safety hazards.
    \item \textbf{Item Organization}: Involves functional items with a defined purpose or designated location (e.g., clothing, toys, utensils, books) that are misplaced or disorganized. These are not debris (non-functional) or contaminants (stains). The task is to return items to their intended spots or arrange them in an orderly manner.
    \item \textbf{Safety Check}: Addresses conditions posing immediate or potential risks of harm (e.g., injury, fire, damage). These override other categories if they introduce danger. The task is to flag or assess such risks.
    \item \textbf{Negative Sample}: the room is tidy, and no outstanding task should be proposed.
\end{itemize}

Examples of snapshots of each category in the Active Task Perception Benchmark are shown in Table~\ref{tab:benchmark_demo}.

The collected snapshots are then processed by QwenVL~\cite{Qwen2.5-VL} and GPT-4o~\cite{gpt4o} through task-specific prompts to generate task proposals and bounding boxes for reference. 
To ensure data quality, we perform a strict manual verification step. Three independent human annotators review each snapshot by consulting the VLM-generated task suggestions and reference boxes, then provide their own labels. A sample is accepted into the benchmark only if all three annotations are similar, i.e., they agree on the task category (debris cleaning, item organization, safety check, or negative) and describe consistent visual evidence for the decision. Samples lacking unanimous agreement are removed from the pool.

During inference, each tested model receives a general prompt to identify all possible tasks in the given snapshot.
For evaluation, as the tested models are not trained to align with the linguistic habits of the ground-truth annotations, we utilize \texttt{ChatGPT-4.1-0414}~\cite{gpt3.5} for evaluating open-ended predictions. The predictions are given a score of $0$, $0.5$, or $1$ point based on correctness and clarity.

\begin{table*}[htbp]
    \centering
    \caption{Zeroshot performance on our Active Task Perception Benchmark. The best results among different models are highlighted in bold, while the second-best results are underlined.}
    \label{tab:exp_benchmark}
    \fontsize{8.6}{11}\selectfont
    
    \begin{tabular}{c|cccc|cc|c}
            \toprule
            \multirow{2}{*}{\textbf{Method}} & \multicolumn{4}{c|}{\textbf{Positive Samples}} & \multicolumn{2}{c|}{\textbf{Negative Samples}} & \textbf{Overall} \\
            
             & \textbf{Clean Debris} & \textbf{Organize Item} & {\textbf{Safety Check}} & {\textbf{Avg. Acc.}} & {\textbf{\# Proposals}} & {\textbf{Acc.}} & {\textbf{Avg. Acc.}}  \\
            \midrule
            \makecell{QwenVL-max~\cite{Qwen2.5-VL}} & \textbf{44.47\%} & 38.22\% & \underline{19.21\%} & \underline{33.97\%} & 0.31 & 81.48\% & \textbf{45.85\%} \\
            \makecell{GPT-o4-mini-0516~\cite{gpto3-o4-mini}} & 36.95\% & \textbf{45.25\%} & 16.21\% & 32.80\% & 0.87 & 57.41\% & \underline{38.95\%}\\
            \makecell{Qwen2.5VL-7B~\cite{Qwen2.5-VL}} & 12.71\% & 4.91\% & 3.45\% & 7.02\% & \textbf{0.00} & \textbf{100.00\%} & 27.76\%\\
            \makecell{MM-Eureka-Qwen-7B~\cite{mmeureka}} & 13.04\% & 6.92\% & 4.35\% & 8.10\% & \underline{0.02} & \underline{98.15\%} & 30.61\%\\
            \makecell{Cosmos-Reason1-7B~\cite{cosmos-reason1}} & \underline{39.13\%} & \underline{41.97\%} & \textbf{24.32}\% & \textbf{35.14\%}  & 1.96 & 31.48\% & 34.23\% \\
            \makecell{RoboBrain-7B~\cite{robobrain1}} & {24.46\%} & {39.32\%} & 9.75\% & {24.51\%} & 1.94  & 11.11\% & 21.11\%\\
            \makecell{RoboBrain2.0-7B~\cite{robobrain2}} & 9.25\% & 8.11\% & 7.76\% & 8.37\% & 0.04 & 96.30\% & 30.35\% \\
            \bottomrule
    \end{tabular}
\end{table*}

\begin{table*}[htbp]
    \centering
    \caption{Performance of the $\mathcal P^3$ framework in real-world deployments. Results demonstrate that the $\mathcal P^3$ framework can effectively perform omni-task perception, seamless plug-and-play tool integration, and effective competing-task planning and coordination across diverse task scenarios. $^*$ indicates that the scheduled task does not need to be completed and \ding{55} denotes that the group of tasks cannot be finished.} \label{tab:real-world}
    \begin{tabular}{cccccccc|c|cc}
        \toprule
            \multirow{2}{*}{Task No.}  & \multicolumn{3}{c}{Task Type} & \multicolumn{4}{c}{Tool Usage} &  RoboOS$^*$ & \multicolumn{2}{c}{$\mathcal P^3$} \\
            \cmidrule(lr){2-4} \cmidrule(lr){5-8}   \cmidrule(lr){9-11}
            ~ & Passive & Active & Scheduled & VLN & VLA & loT & Web & GPT-o4-mini & GPT-o4-mini & Qwen-max  \\
            \midrule
            \rowcolor{gray!20} \multicolumn{11}{l}{\textbf{Tool Integration and Execution}} \\
            \midrule
            1 & ~ & \checkmark & ~ & ~ & \checkmark & ~ & ~ & \ding{55} &85.71 & 78.57  \\ 
            2 & \checkmark & \checkmark & ~ & ~ & ~ & \checkmark & ~ & \ding{55} &100.00 & 100.00  \\ 
            3 & ~ & ~ & \checkmark & \checkmark & ~ & \checkmark & ~ & \ding{55} &78.57 & 71.43  \\ 
            4 & \checkmark & ~ & ~ & ~ & ~ & ~ & \checkmark & 100.00 &100.00 & 100.00  \\ 
            5 & ~ & ~ & \checkmark & \checkmark & \checkmark & ~ & ~ & 92.86 &100.00 & 92.86  \\ 
            6 & \checkmark & ~ & ~ & \checkmark & \checkmark & ~ & \checkmark & 78.57 &78.57 & 85.71 \\ 
        \midrule
        \rowcolor{gray!20} \multicolumn{11}{l}{\textbf{Competing-task Planning and Coordination}} \\
        \midrule
        7 & \checkmark & ~ & \checkmark & \checkmark & \checkmark & \checkmark & \checkmark & \ding{55} &92.86 & 92.86  \\ 
        8 & \checkmark & ~ & \checkmark & \checkmark & \checkmark & \checkmark & \checkmark & \ding{55} &92.86 & 85.71  \\ 
        9 & \checkmark & \checkmark & \checkmark & \checkmark & \checkmark & \checkmark & \checkmark  & \ding{55} &64.29 & 71.43  \\ 
        10 & \checkmark & \checkmark & \checkmark & \checkmark & \checkmark & \checkmark & \checkmark & \ding{55} &57.14 & 71.43 \\
        11 & \checkmark & \checkmark & \checkmark & \checkmark & \checkmark & \checkmark & \checkmark & \ding{55} &35.71 & 42.86 \\
    \bottomrule
    \end{tabular}
\end{table*}

\begin{figure*}[tp]
    \centering
    \includegraphics[width=0.9\textwidth]{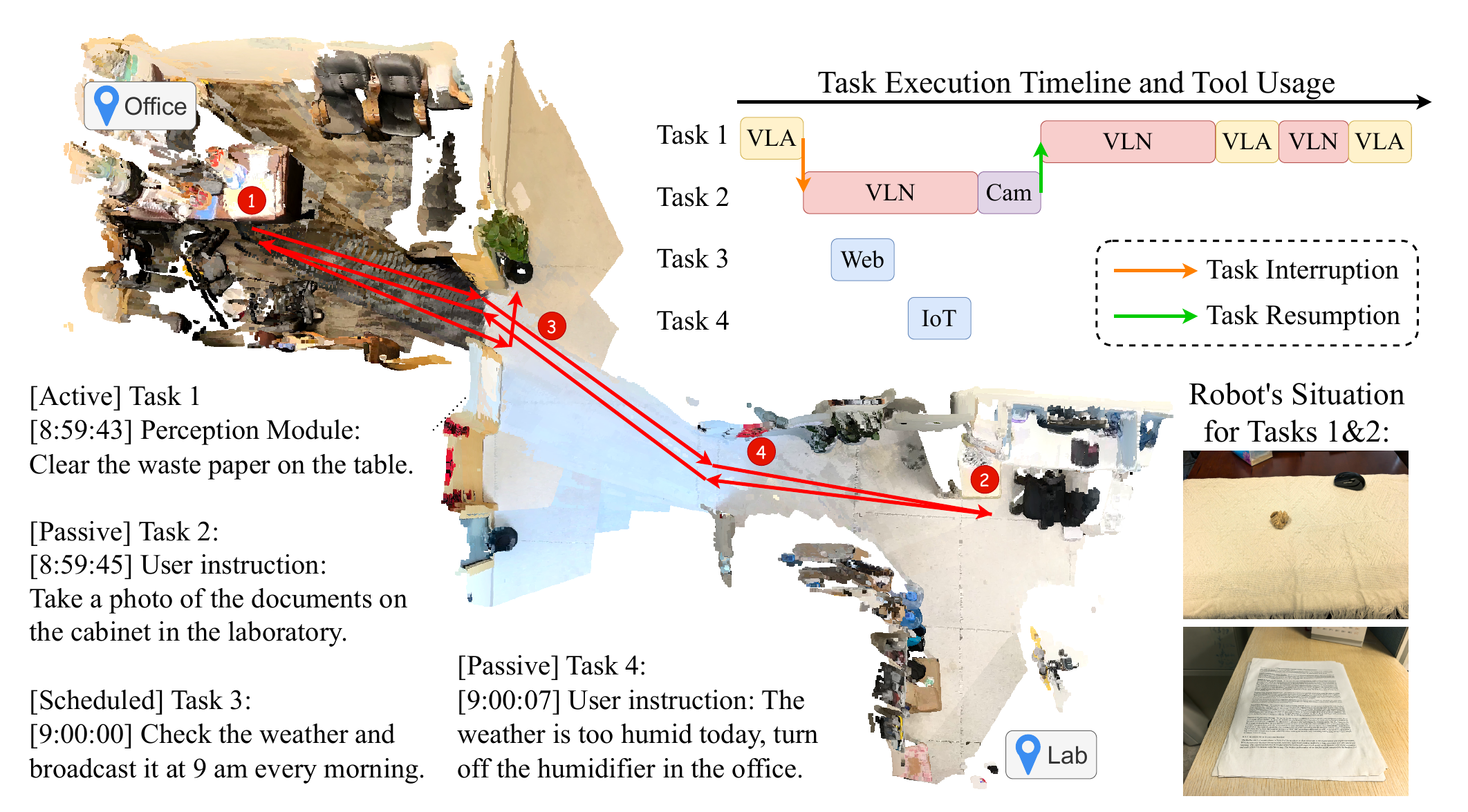}
    \caption{Execution illustration of Task 9 from Table~\ref{tab:real-world} where a spatial-temporal task map with four tasks was executed across different rooms. The arrows trace the robot’s navigation trajectory, while the timeline above visualizes task execution flow and tool usage. Interruptions (orange) and resumptions (green) reflect dynamic task management.}
    \label{fig:example}
\end{figure*}

Results in Table~\ref{tab:exp_benchmark} show that QwenVL-max~\cite{Qwen2.5-VL} strikes the best balance between positive samples and negative samples, achieving an overall accuracy of 45.85\%, making it the most suitable option for the perception module. For positive samples, Cosmos-Reason1-7B~\cite{cosmos-reason1} and QwenVL-max~\cite{Qwen2.5-VL} achieve the highest accuracies of 35.14\% and 33.97\%, respectively. However, for tidy scenes, they always try to identify flaws within them, resulting in unnecessary proposals. On the other side, models like Qwen2.5VL-7B~\cite{Qwen2.5-VL}, MM-Eureka-Qwen-7B~\cite{mmeureka}, and RoboBrain2.0~\cite{robobrain2} prefer to generate fewer proposals, yielding fewer active actions in real-world deployments. Finally, as the $\mathcal P^3$ framework does not rely on a specific VLM for the perception module, users can adjust the preference for the number of proposals by utilizing different baseline VLMs or guiding through additional prompts.

\subsection{Real-world Deployment Evaluation}

To thoroughly validate the versatility and robustness of the $\mathcal P^3$ framework, we design a suite of $11$ real-world embodied tasks for robotic execution. These tasks are categorized into independent tasks and competing tasks. The independent tasks are designed to assess the framework's ability to utilize diverse tools and can be executed independently without interference from other tasks. In contrast, competing tasks involve executing multiple tasks on limited resources, which is used to evaluate the framework's capacity to perform multi-task planning.

\noindent \textbf{Independent Tasks.}
To evaluate $\mathcal P^3$'s tool integration and execution capabilities, we designed six distinct tasks, encompassing various types of tool agents, including navigation, manipulation, IoT, and web modules. These tasks reflect real-world scenarios in which the robot must collaborate with multiple tools to achieve its goals. 
The task definitions and expected execution flows are summarized in Tables~\ref{tab:independent} and~\ref{tab:compating}.
During the experiment, each task is given sufficient time for execution. Empirically, all successful attempts can be accomplished within one minute, which is acceptable in real-world applications.

Results in Table~\ref{tab:real-world} demonstrate that the $\mathcal P^3$ framework exhibited robust performance across these tasks, showcasing its versatility in integrating diverse tools. For IoT-related tasks (e.g., Task~2, turning off a humidifier), the framework achieved high success rates, indicating proficiency in interacting with smart devices. Similarly, Task~1 (autonomously picking up scrap paper) was completed without explicit human instruction, revealing $\mathcal P^3$’s ability to \emph{actively perceive} and generate actionable goals from environmental changes rather than waiting for user commands. In contrast, RoboOS lacks such an active perception mechanism and instead follows a purely \emph{passive execution paradigm}, where the agent performs only predefined user-issued instructions. As a result, RoboOS is unable to detect spontaneous opportunities for action—such as identifying scattered debris or recognizing newly emerged cleaning tasks—and thus cannot initiate behaviors beyond the instructed sequence. 
By continuously monitoring the environment through its perception module, $\mathcal P^3$ can autonomously propose and execute relevant tasks, which greatly enhances its adaptability in dynamic settings. This capability becomes particularly evident in more complex multi-step operations, such as Task~6 (``Move the cardboard box and place an order for delivery''), where $\mathcal P^3$ effectively combines navigation, manipulation, and web API calls. Together, these results underline that $\mathcal P^3$ not only integrates diverse tools but also extends the behavioral autonomy of embodied agents beyond the passive instruction-following paradigm characteristic of RoboOS.

While our $\mathcal P^3$ framework performed well across various tasks, certain challenges remain. On one hand, due to the imperfect performance 
of LLMs for the perception and planning modules, $\mathcal P^3$ cannot always generate optimal answers. On the other hand, the robots' unstable hardware has also caused multiple failures during the experiment (e.g., unstable network connections and overheating protections). Thus, with improvements of individual modules, $\mathcal P^3$ can further enhance its adaptability and effectiveness in more complex real-world scenarios.

\begin{table*}[!ht]
    \centering
    \caption{Independent Task List}
    \label{tab:independent}
    \renewcommand{\arraystretch}{1.6} 
    \setlength{\tabcolsep}{6pt}
    \begin{tabular}{c|p{4.5cm}p{9cm}}
    \hline
        Task ID & Task Description & Expected Execution Flow  \\ \hline
        1 & Autonomously pick up waste paper on the table & 1. The Perception Module actively detects waste paper on the table. 2. The Manipulation Agent (VLA) is deployed to grasp the waste paper. 3. The Perception Module confirms task completion.  \\ \hline
        2 & Instruction ``It's too humid in the room''  & 1. User instruction triggers a passive task. 2. IoT tools are used to turn off the humidifier.  \\ \hline
        3 & Instruction ``Check the office for people after 1 minute; turn off the lights if no one is present'' & 1. A scheduled task is triggered by the timer after 1 minute. 2. The Navigation Agent (VLN) navigates to the office. 3. The Perception Module detects human presence. 4. If no one is present, IoT tools turn off the lights. 5. The Perception Module confirms completion.  \\ \hline
        4 & Instruction ``I plan to go out, provide dressing suggestions'' & 1. User instruction triggers a passive task. 2. Web tools retrieve real-time weather information. 3. Dressing suggestions are generated based on the weather.  \\ \hline
        5 & Instruction ``Bring the medicine box to the office every day at 11:00 a.m.'' & 1. A scheduled task is triggered by the timer daily at 11:00 a.m. 2. The Manipulation Agent (VLA) picks up the box. 3. The Navigation Agent (VLN) guides to the office for placement. 4. The Perception Module confirms placement.  \\ \hline
        6 & Instruction ``Move the small cardboard box on the lab table to the lab cabinet and place an order for delivery'' & 1. User instruction triggers a passive task. 2. The Navigation Agent (VLN) navigates to the lab table. 3. The Manipulation Agent (VLA) picks up the cardboard box. 4. VLN goes to the lab cabinet and VLA puts the box in the cabinet. 6. Web tools are used to place a courier order. \\ \hline
    \end{tabular}
\end{table*}

\begin{table*}[!ht]
    \centering
    \caption{Competing Task List}
    \label{tab:compating}
    \renewcommand{\arraystretch}{1.6} 
    \setlength{\tabcolsep}{6pt}
\begin{tabular}{c|p{5cm}p{9cm}}
    \hline
        Task ID & Task Description & Expected Execution Flow  \\ \hline
        1 & Robot is in an orderly lab; first, engage in conversation to ask for its function, then ask it to play music after 10 seconds; next, instruct it to go to the office table and retrieve the medicine box. & 1. User interacts with the robot to ask about its function. 2. The Web Agent plays music after 10 seconds. 3. The Navigation Agent (VLN) guides the robot to the office. 4. The Manipulation Agent (VLA) picks up the medicine box. 5. The Perception Module confirms task completion. \\ \hline
        2 & Robot is in an orderly lab; first, engage in conversation to ask for its function; then, instruct it to go to the office table and retrieve the medicine box, interrupt navigation mid-way to turn on the office lights. & 1. User interacts with the robot to ask about its function. 2. The Navigation Agent (VLN) starts to guide the robot to the office. 3. Midway, the Navigation Agent is interrupted, and the Perception Module detects no lights in the office. 4. IoT tools turn on the office lights. 5. The Navigation Agent (VLN) returns to the office to retrieve the medicine box. 6. The Manipulation Agent (VLA) retrieves the medicine box. 7. The Perception Module confirms task completion. \\ \hline
        3 & Robot is in the office; discovers an active task in the scene, picks up waste paper from the desk; suddenly interrupted to go to the lab cabinet to take a photo of a document; while en route, triggers a scheduled task to check the weather at 9 a.m. and broadcast the result; then, while taking the document photo, user sends an instruction to turn off the humidifier in the office. & 1. The Perception Module detects waste paper on the office desk. 2. The Manipulation Agent (VLA) picks up the waste paper. 3. A sudden interruption triggers a new task, and the Navigation Agent (VLN) guides the robot to the lab cabinet. 4. The Manipulation Agent (VLA) takes a photo of the document on the cabinet. 5. Mid-route, a scheduled task triggers: The Weather Check Agent retrieves and broadcasts weather information. 6. The user sends an instruction to turn off the office humidifier via IoT tools. 7. The Perception Module confirms task completion.  \\ \hline
        4 & Robot is in the lab; first, engage in conversation to ask for its function; then, instruct it to go to the office table and retrieve the medicine box; midway, interrupt navigation and instruct it to place a package on the desk in the office; the final task is to discover an empty bottle on the office desk and throw it into the trash can. & 1. User interacts with the robot to ask about its function. 2. The Navigation Agent (VLN) starts to guide the robot to the office. 3. Midway, the Navigation Agent is interrupted. 4. The Manipulation Agent (VLA) places a package on the desk in the office. 5. The Navigation Agent (VLN) resumes and completes the task of retrieving the medicine box. 6. The Perception Module detects an empty bottle on the office desk. 7. The Manipulation Agent (VLA) grabs the discarded bottle. 8. The Perception Module confirms task completion. \\ \hline
        5 & Robot is in the office asked about the weather; suddenly interrupted to go to the lab cabinet to take a photo of a document; in the middle of the route, it discovers knocked-over cup on the cabinet and picks it up, triggering a scheduled task for weather check at 9 a.m. to broadcast, and user sends an instruction to turn off the humidifier in the office. & 1. User asks the robot about the weather, triggering a task. 2. The Navigation Agent (VLN) starts guiding the robot to the lab cabinet. 3. Mid-route, the robot detects a knocked-over cup on the cabinet. 4. The Manipulation Agent (VLA) picks up the knocked-over cup. 5. The Weather Check Agent triggers and broadcasts the weather forecast. 6. The user sends an instruction to turn off the office humidifier via IoT tools. 7. The Manipulation Agent (VLA) takes a photo of the document on the cabinet. 8. The Perception Module confirms task completion. \\ \hline
    \end{tabular}
\end{table*}

\noindent \textbf{Competing Tasks.}
In the next set of experiments, we evaluated the $\mathcal P^3$ framework’s ability to handle tasks competing for limited resources, which poses a harder challenge by incorporating individual task execution with complex task planning requirements. While independent tasks stem from a single command and can be broken down into sub-tasks without resource conflict, competing tasks involve both sequential and concurrent execution with interdependent priorities.
For example, in Task 11, the robot receives multiple instructions, each requiring decomposition into smaller sub-tasks and careful prioritization. The challenge becomes harder when new instructions arrive during ongoing execution. As illustrated in Fig.~\ref{fig:example}, the robot must switch from a suspended task to a higher-priority instruction, complete the urgent task, and then resume the original task. This process depends on the dynamic scheduler, which maintains task state and determines the proper execution order. Task details and execution flows are summarized in Tables~\ref{tab:independent} and~\ref{tab:compating}.

Table~\ref{tab:real-world} highlights the $\mathcal P^3$ framework's success in managing competing-task execution. 
Tasks 7 and 8 showed high success rates, attributed to the clean and simple environment that minimized irrelevant task detections. However, Tasks 9, 10, 
and 11 presented greater challenges, as active tasks were introduced mid-execution, requiring real-time adjustments. Task 11, in particular, involved picking up a knocked-over cup actively, requiring the robot to adapt its actions and modify its plan on the fly, resulting in a slight drop in performance due to the increased complexity.

\begin{figure*}[!p]
    \centering
    \includegraphics[width=0.9\textwidth]{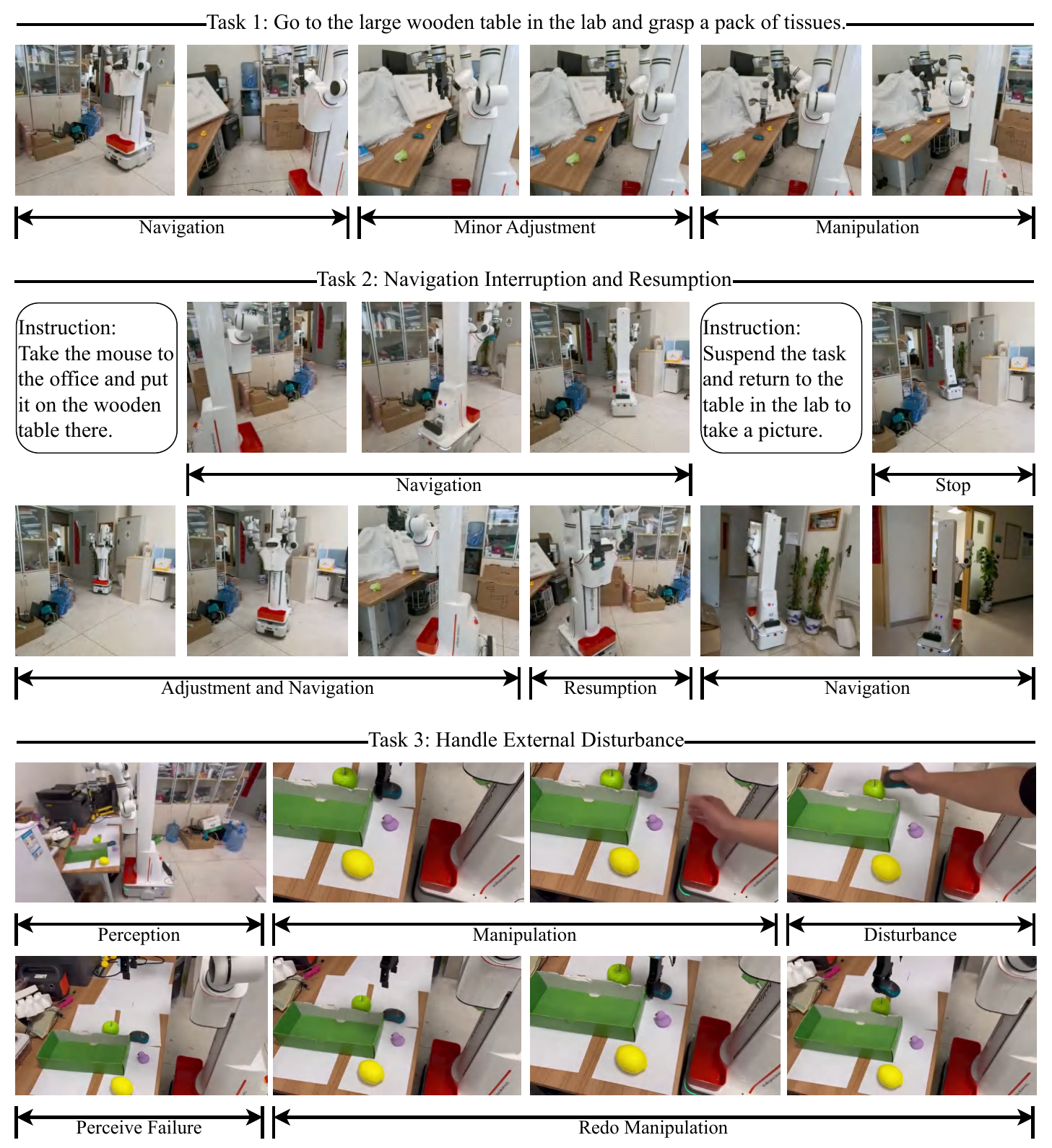}
    \caption{Qualitative results for real-world deployments. In the first task, the robot is given the instruction to ``go to the large wooden table in the lab and grasp a pack of tissues.'' When the robot approaches the table, it perceives that it is still some distance away from the table and performs a minor adjustment by moving forward. This task demonstrates that our $\mathcal P^3$ framework can help the robot adjust flexibly according to the complex environment. In Task 2, the robot is given an instruction to take the mouse to the office and then given another instruction with higher priority to suspend the previous task and return to the table to take a picture. When the robot receives the second instruction, the planner module in the $\mathcal P^3$ framework interrupts the low-priority task and temporarily puts it back in the task memory. When the latter task is finished, the suspended task is resumed automatically, showing strong capability in handling multiple tasks. Finally, in the last task, when the robot is grabbing the mouse, an external disturbance (human activity) happens. As the perception module detects the unexpected outcome, it pauses the original workflow and reverts to the grabbing process, verifying that our $\mathcal P^3$ framework is robust to external disturbance.}
    \label{fig:qualitative}
\end{figure*}

Nonetheless, our $\mathcal P^3$ framework demonstrated robust competing-task management throughout. Its dynamic scheduler and task memory allowed robots to efficiently manage tasks even in the presence of interruptions, unexpected events, or newly introduced instructions. 
Additional qualitative cases in Fig.~\ref{fig:qualitative} visualize navigation adjustment, interruption and resumption, and recovery from external disturbance.
By contrast, RoboOS demonstrated fundamental limitations under the same settings. Without a persistent task representation or dependency management, RoboOS could only execute instructions in a strictly linear order. When interruptions occurred or new commands arrived mid-execution, it either abandoned the previous task or failed to initiate the new one, resulting in incomplete sequences.
\section{Conclusion}

In this work, we introduce $\mathcal P^3$, a unified framework designed to advance embodied agents toward versatile and scalable real-world deployment. By introducing an active perception module capable of autonomously identifying tasks, our approach eliminates the dependency on restrictive tool feedback. This architectural innovation enables truly plug-any-tool capability, allowing seamless integration with any controllable device. We further pair this with a dynamic multi-task scheduler that can effectively handle dependencies and interruptions, better adapting to practical requirements. Extensive real-world experiments confirm that our $\mathcal P^3$ framework successfully bridges the gap between benchmarks and practical deployment, delivering a more versatile and adaptable embodied agent. It also provides a foundation for scalable embodied data collection.

\clearpage
\bibliographystyle{plainnat}
\bibliography{references}

\end{document}